\documentclass[12pt]{elsarticle}
\pdfoutput=1 % for arXiv

\RequirePackage{snapshot}

\biboptions{sort&compress}

\usepackage{url}
\usepackage{graphicx}

\usepackage{pdfpages} % to include pdf graphics
\usepackage{amsmath,amssymb} % define this before the line numbering.
\usepackage[american]{babel} % language selection, hyphenation, ...
\usepackage{amsmath} % assumes amsmath package installed
\usepackage{amssymb}  % assumes amsmath package installed
\usepackage{mathrsfs}
\usepackage{dsfont}
\usepackage{esint}
\usepackage[T1]{fontenc}
\usepackage[babel]{csquotes} % \enquote{...}
\usepackage{xcolor} % provide colors (at least for the writing process)
\usepackage{pdflscape} % Package pdfpages Warning: You are using `thumbpdf.sty' but not `pdflscape.sty'. Please, insert `pdflscape.sty' in your document to get a pleasant PDF document with thumbs.
\usepackage{url}
\usepackage{subfigure}
\usepackage{siunitx}
\usepackage{hyperref}

\newcommand{\ul}[1]{\underline{#1}}

\journal{Image and Vision Computing}

\begin{document}

\begin{frontmatter}



\title{On the Distribution of Salient Objects in Web Images \\ and its Influence on Salient Object Detection}

\author{Boris Schauerte\fnref{kit}}
\cortext[cor1]{Corresponding author}
\ead{boris.schauerte@kit.edu}
% \ead[url]{https://cvhci.anthropomatik.kit.edu/~bschauer/}
\author{Rainer Stiefelhagen\fnref{kit}}%\fnref{kit,szs,pui}}
\ead{rainer.stiefelhagen@kit.edu}
\address[kit]{Institute for Anthropomatics, Karlsruhe Institute of Technology (KIT), \\ 76131 Karlsruhe, Germany}

\begin{abstract}
In recent years it has become apparent that a Gaussian center bias can serve as an important prior for visual saliency detection, which has been demonstrated for predicting human eye fixations \cite{judd2009learning,yang2010what,borji2012probabilistic} and salient object detection \cite{jiang2011automatic}.
Tseng et al. have shown that the photographer's tendency to place interesting objects in the center is a likely cause for the center bias of eye fixations \cite{tseng2009quantifying}.
We investigate the influence of the photographer's center bias on salient object detection, extending our previous work \cite{schauerte2013distribution}.
We show that the centroid locations of salient objects in photographs of Achanta and Liu's data set \cite{liu2007learning,achanta2009frequency-tuned} in fact correlate strongly with a Gaussian model.
This is an important insight, because it provides an empirical motivation and justification for the integration of such a center bias in salient object detection algorithms and helps to understand why Gaussian models are so effective.
To assess the influence of the center bias on salient object detection, 
we integrate an explicit Gaussian center bias model into two state-of-the-art salient object detection algorithms \cite{cheng2011global,achanta2010saliency}.
This way, first, we quantify the influence of the Gaussian center bias on pixel- and segment-based salient object detection.
Second, we improve the performance in terms of $F_1$ score, $F_\beta$ score, area under the recall-precision curve, area under the receiver operating characteristic curve, and hit-rate on the well-known data set by Achanta and Liu \cite{liu2007learning,achanta2009frequency-tuned}.
Third, by debiasing Cheng et al.'s region contrast model, we exemplarily demonstrate that implicit center biases are partially responsible for the outstanding performance of state-of-the-art algorithms.
Last but not least, as a result of debiasing Cheng et al.'s algorithm, we introduce a non-biased salient object detection method, which is of interest for applications in which the image data is not likely to have a photographer's center bias 
(e.g., image data of surveillance cameras or autonomous robots).
\end{abstract}

\begin{keyword}
% 	68T45  	Machine vision and scene understanding
%	68T99  	None of the above, but in this section
%	68U10  	Image processing


Salient Object Detection \sep Object Distribution \sep Photographer Bias


\MSC 68T45 \sep 68U10

\end{keyword}

\end{frontmatter}

% \linenumbers

\newcommand{\opt}{\operatornamewithlimits{opt}}
\newcommand{\sgn}{\operatornamewithlimits{sgn}}
\newcommand{\argmax}{\operatornamewithlimits{argmax}}
\def\bigdot{\ensuremath{\bullet}}
\def\cel{\ensuremath{^{\circ}C}}
\def\grad{\ensuremath{^\circ}}

\def\Re{\operatorname{Re}}
\def\Im{\operatorname{Im}}
\def\Phase{\operatorname{\Phi}}
\def\Magnitude{\operatorname{Mg}}
\def\FT{\operatorname{\mathrm{FT}}}
\def\STFT{\operatorname{\mathrm{STFT}}}
\def\SPECT{\operatorname{\mathrm{SPEC}}}

\def\DCT{\operatorname{\mathrm{DCT}}}
\def\IDCT{\operatorname{\mathrm{IDCT}}}
\def\QDCT{\operatorname{\mathrm{QDCT}}}
\def\IQDCT{\operatorname{\mathrm{IQDCT}}}

\newcommand{\optionalhoutext}[2]{{#2}}

\def\Deriv{\operatorname{d}}

\newcommand{\footnoteremember}[2]{%
  \footnote{#2}%
  \newcounter{#1}%
  \setcounter{#1}{\value{footnote}}%
}
\newcommand{\footnoterecall}[1]{%
  \footnotemark[\value{#1}]%
}

\newcommand{\card}[1]{\left|{#1}\right|}

\newcommand{\mysubsubsection}[1]{\paragraph{#1}}
\newcommand{\optionalcite}[2]{\cite{#2}}
\newcommand{\optionaltextx}[2]{#2}

\newcommand{\vabs}[1]{\lvert#1\rvert}
\newcommand{\vnorm}[1]{\lVert#1\rVert}

\definecolor{orange}{rgb}{1,0.5,0}

\newcommand{\needsrewrite}[1]{{\color{orange}#1}}
\newcommand{\unfinishedtext}[1]{{\color{red}#1}}
\newcommand{\missingtext}[1]{{\color{red}#1}}
\newcommand{\optionaltext}[1]{{\color{gray}#1}}
\newcommand{\todonote}[1]{{\color{green}#1}}
\newcommand{\optionalfootnote}[1]{\footnote{\color{gray}#1}}

\newcommand{\manufacturername}[1]{{\textup{#1}}}
\newcommand{\productname}[1]{{\textup{#1}}}

\newcommand{\confer}{see}

\section{Introduction}\label{sec:introduction}

\begin{figure}[t]
  \centering
  \subfigure[Example images]{
  \includegraphics[height=2cm]{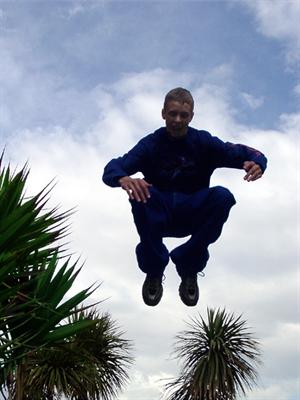}\hspace{0.05cm}
  \includegraphics[height=2cm]{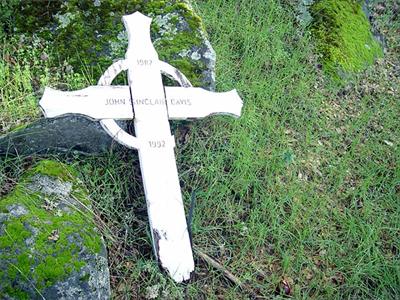}\hspace{0.05cm}
  \includegraphics[height=2cm]{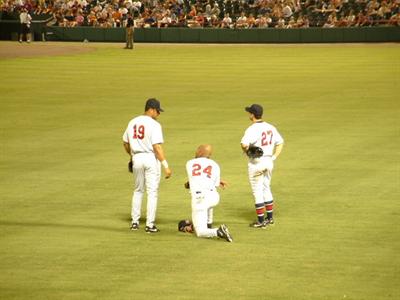}\hspace{0.05cm}
  \includegraphics[height=2cm]{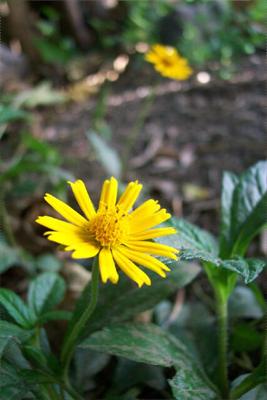}\label{fig:introduction:sub:examples}}\hspace{1em}
  \subfigure[Centroid scatter plot]{
  \includegraphics[height=2.0cm,width=3.8cm]{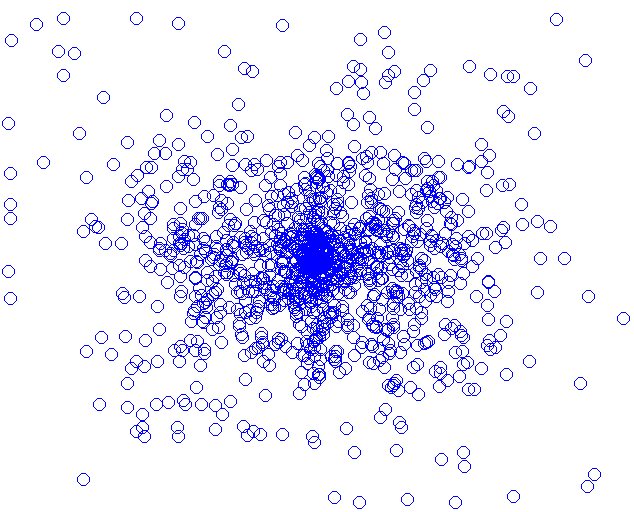}\label{fig:introduction:sub:scatter}}\\
  \subfigure[Example segmentation masks]{
  \includegraphics[height=2cm]{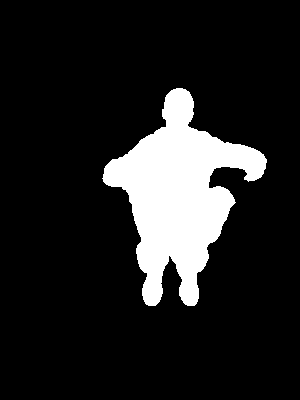}\hspace{0.05cm}
  \includegraphics[height=2cm]{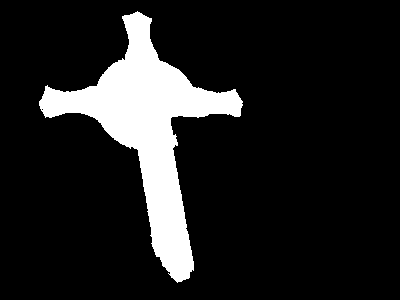}\hspace{0.05cm}
  \includegraphics[height=2cm]{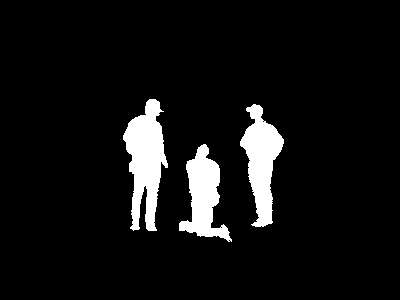}\hspace{0.05cm}
  \includegraphics[height=2cm]{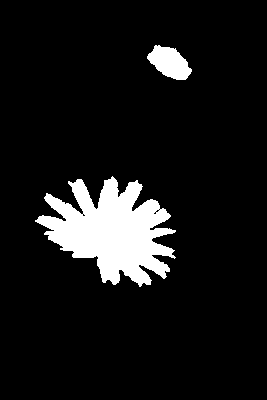}\label{fig:introduction:sub:masks}}\hspace{1em}
  \subfigure[Mean segment mask]{
  \includegraphics[height=2.0cm,width=3.8cm]{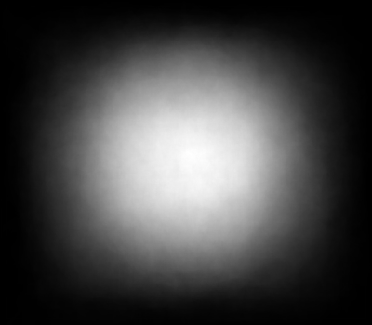}\label{fig:introduction:sub:meanmask}
  }
  \caption{Illustration of the Achanta/Liu data set: example images \ref{fig:introduction:sub:examples}, the corresponding segmentation masks \ref{fig:introduction:sub:masks}, the mean over all segmentation masks \ref{fig:introduction:sub:meanmask}, and the scatter plot of the centroid locations across all images \ref{fig:introduction:sub:scatter}.}
  \label{fig:introduction:meanlocation}
\end{figure}

Among other influences such as task-specific factors, human attention is attracted to salient stimuli.  
In this context, saliency describes the subjective, perceptual quality that lets some items in the world stand out from their neighbors and immediately grab our attention.
Accordingly, the goal of visual saliency detection is to determine what parts of an image are likely to grab the human attention.
The task of \enquote{traditional} {\it visual saliency detection} is to predict where human observers look when presented with a scene, which can be recorded using eye tracking equipment (e.g., \cite{einhaeuser2008objects,yang2010what,judd2009learning,schauerte2012quaternion-based}).
Liu et al. adapted the traditional definition of visual saliency by incorporating the high level concept of a salient object into the process of visual attention computation \cite{liu2007learning}.
Here, a {\it salient object} is defined as being the object in an image that attracts most of the user's interest such as, for example, the man, the cross, the baseball players and the flowers in Fig.~\ref{fig:introduction:sub:examples} (left-to-right, resp.).
Accordingly, Liu et al. \cite{liu2007learning} defined the task of {\it salient object detection} as the binary labeling problem of separating the salient object from the background.
Thus, in contrast to traditional visual saliency detection, salient object detection does not just comprise of the task to calculate the saliency of image regions, but it also incorporates the task to determine and segment the most salient object in the image.
Here, it is important to note that the selection of a salient object happens consciously by the user whereas the gaze trajectories that are recorded using eye trackers are the result of mostly unconscious processes.
Consequently, also taking into account that salient objects attract the human gaze (see, e.g., \cite{einhaeuser2008objects}), salient object detection and predicting where people look are very closely related yet substantially different tasks.

The photographer's center bias, i.e. the natural tendency of photographers to place the objects of interest near the center of their composition in order to enhance their focus and size relative to the background (see \cite{tseng2009quantifying}), has been identified as one cause for the often reported center bias in eye-tracking data during eye-gaze studies \cite{reinagel1999natural,parkhurst2003scene,tatler2007central}\footnote{Here, it is important to note that Tseng et al. -- due to their methodology -- did not investigate the exact spatial distribution of the objects that attract the gaze. They hired five persons who provided subjective scores from 1 to 5 in terms of how interesting things were biased toward the image center \cite{tseng2009quantifying}.}.
As a consequence, the integration of a center bias has become an increasingly important aspect in visual saliency models that focus on gaze prediction (e.g., \cite{yang2010what,judd2009learning,borji2012probabilistic}).
In contrast, most recently proposed salient object detection algorithms do not incorporate an explicit model of the photographer's center bias (see, e.g., \cite{achanta2009frequency-tuned,achanta2010saliency,klein2011center-surround,cheng2011global}).
A notable exception and closely related to our work is the work by Jiang et al. \cite{jiang2011automatic}, in which one of the three main criteria that characterize a salient object is that \enquote{it is most probably placed near the center of the image} \cite{jiang2011automatic}.
The authors justify this characterization with the \enquote{rule of thirds}, which is one of the most well-known principles of photographic composition (see, e.g., \cite{luo2008photo}), and use a Gaussian distance metric as a model.
We
go beyond following the rule of third and show that the distribution of the objects' centroids correlates strongly positively with a 2-dimensional Gaussian distribution. 
This means nothing less than that we 
provide a strong empirical justification for integrating Gaussian center bias models into salient object detection algorithms.
To demonstrate the importance,
we adapt two state-of-the-art salient object detection methods to quantify the influence of the photographer's center bias on salient object detection.

The contribution of this paper is twofold:
First, we use the salient object data set by Achanta et al. \cite{achanta2009frequency-tuned} to investigate the spatial distribution of salient objects in images. 
This way, in Sec.~\ref{sec:hypotheses}, we show that it is likely that salient objects in photographs are distributed around the image center in such a way that the radii are half-Gaussian distributed and the angles are uniformly distributed.
Second, in Sec.~\ref{sec:evaluation:salobj}, we explicitly integrate Gaussian center bias models in two recently proposed salient object detection methods: 
The pixel-based maximum symmetric surround salient object detection by Achanta et al. \cite{achanta2010saliency} and the segment-based region contrast method by Cheng et al. \cite{cheng2011global}.
In order to measure the influence, we use the following evaluation measures: The maximum $F_1$ score, the maximum $F_{\beta}$ score with $\beta=\sqrt{0.3}$ \cite{achanta2009frequency-tuned}, the area-under-curve of the precision-recall curve, the AUC of the receiver operating characteristic (ROC AUC), and the hit-rate.
In summary, the integration of the center bias model increases the ROC AUC by \SI{2}{\percent} and the performance with respect to all remaining measures by roughly \SI{5}{\percent}.
Thus, we further advance the state-of-the-art of pixel-based as well as segment-based salient object detection.
By modifying Cheng et al.'s region contrast model \cite{cheng2011global}, first, we obtained a non-biased salient object detection algorithm that is based on region contrast and, second, we exemplarily demonstrate that implicit center biases can already be found in well-performing, state-of-the-art salient object detection algorithms and substantially influence the performance.
This is important to consider when comparing and selecting algorithms for applications in which the data is not necessarily biased towards the center.

The remainder of this paper is organized as follows:
In Sec.~\ref{sec:relwork}, we provide an overview of related work. 
Subsequently, in Sec.~\ref{sec:hypotheses}, we introduce and investigate our hypotheses about the spatial distribution of salient objects.
Then, in Sec.~\ref{sec:evaluation:salobj}, we integrate our hypotheses into two recently proposed salient object detection methods and evaluate the influence on the salient object detection performance.
We conclude with a short summary and discussion in Sec.~\ref{sec:conclusion}.
Furthermore, please feel free to check the supplemental material for additional information such as, e.g., further evaluation results.

\section{Related Work}\label{sec:relwork}

We focus on the most recent related work that addresses bottom-up saliency detection with an emphasis on salient object detection (see, e.g., \cite{tsotsos2011computational} for a more general overview of computational attention models).
Such methods may be biologically motivated, or purely computational, or involve both aspects. 
In 2009, Achanta et al. \cite{achanta2009frequency-tuned,achanta2010saliency} introduced a salient object detection approach that basically relies on the difference of pixels to the average color and intensity value.
In order to evaluate their approach, they selected a sub-set of 1000 images of the image data set that was collected from the web by Liu et al. \cite{liu2007learning} and calculated segmentation masks of the salient objects that were marked by 9 participants using (rough) rectangle annotations \cite{liu2007learning}.
Please note that this procedure also means that during the manual data set annotation the selection of the salient object happens mostly conscious whereas gaze trajectories that are recorded using eye trackers are a result of a mostly unconscious process.
Since it was created, the salient object data set by Achanta et al. serves as reference data set to evaluate methods for salient object detection (see, e.g., \cite{achanta2009frequency-tuned,achanta2010saliency,klein2011center-surround,cheng2011global}).
Liu et al. \cite{liu2007learning} and Alexe et al. \cite{alexe2010what} approach salient object detection using machine learning. 
To this end, Liu et al. \cite{liu2007learning} combine multi-scale contrast, center-surround histograms, and color spatial-distributions with conditional random fields.
Similarly, Alexe et al. \cite{alexe2010what} combine multi-scale saliency, color contrast, edge density, and superpixels in a Bayesian framework.
Closely related to Bayesian surprise \cite{itti2006bayesian}, Klein et al. \cite{klein2011center-surround} use the Kullback-Leibler Divergence of the center and surround image patch histograms to calculate the saliency.
Cheng et al. \cite{cheng2011global} use segmentation to define a regional contrast-based method, which simultaneously evaluates global contrast differences and spatial coherence.
Here, we can differentiate between algorithms that rely on segmentation-based (e.g., \cite{cheng2011global,alexe2010what}) and pixel-based contrast measures (e.g., \cite{achanta2009frequency-tuned,achanta2010saliency,klein2011center-surround}).
Closely related to our work on the quantitative influence of the center bias on salient object detection is the work by Jiang et al. \cite{jiang2011automatic} and most recently Borji et al. \cite{borji2012salient}.
In Jiang et al.'s work \cite{jiang2011automatic} one of the main criteria that characterize a salient object is that \enquote{it is most probably placed near the center of the image}, which is justified with the \enquote{rule of thirds}.
Most recently, Borji et al. \cite{borji2012salient} evaluated several salient object detection models and also performed tests with an additive Gaussian center bias and conclude that the resulting \enquote{change in accuracy is not significant and does not alter model rankings}.
But, this neglects the possibility that well-performing models already have an integrated, implicit center bias, which -- as one part of our work -- we demonstrate exemplarily to be the case for Cheng et al.'s region contrast algorithm \cite{cheng2011global}.
Furthermore, there exist several approaches that explicitly integrate a center bias, but do not provide a quantitative evaluation of its influence nor an empirical justification of the chosen model (e.g., \cite{scharfenberger2013statistical}).
In this paper, we adapt the pixel-based method by Achanta et al. \cite{achanta2010saliency} and the segmentation-based method by Cheng et al. \cite{cheng2011global} to incorporate a model of the photographer-related center bias and quantify the influence of the center bias on the performance.
Furthermore, Borji et al. \cite{borji2012salient} do not provide an empirical justification why a Gaussian distribution is an appropriate center bias model, which is another part of the work described in this paper.

It has been observed in several studies that the visual attention of human participants in natural scenes is biased toward the center of static images and videos (see, e.g., \cite{busswell1935how,tatler2007central,parkhurst2003scene}).
One possible bottom-up cause of the bias is intrinsic bottom-up visual saliency as predicted by computational saliency models.
One possible top-down cause of the center bias is known as photographer bias (see, e.g., \cite{reinagel1999natural,parkhurst2003scene,tatler2007central}), which describes the natural tendency of photographers to place objects of interest near the center of their composition.
In fact, what the photographer considers interesting may also be highly bottom-up salient.
Additionally, the photographer bias may lead to a viewing strategy bias \cite{parkhurst2002modeling}, which means that viewers may orient their attention more often toward the center of the scene, because they expect salient or interesting objects to be placed there.
Thus, since in natural images and videos the distribution of objects of interest and thus saliency is usually biased toward the center, it is often unclear how much the saliency actually contributes in guiding attention. 
It is possible that people look at the center for reasons other than saliency, but their gaze happens to fall on salient locations.
Therefore, this center bias may result in overestimating the influence of saliency computed by the model and contaminate the evaluation of how visual saliency may guide orienting behavior.
Recently, Tseng et al. \cite{tseng2009quantifying} were able to demonstrate quantitatively that center bias is correlated strongly with photographer bias and is influenced by viewing strategy at scene onset.
Furthermore, e.g., they were able to show that motor bias had no effect.
However, they did not evaluate and computationally model how specifically the objects that attract the gaze are distributed spatially in the image. Instead, Tseng et al. hired five naive participants to provide subjective scores from 1 to 5 in terms of how interesting things were biased toward the image center \cite{tseng2009quantifying}.
In this paper, we use the data set by Achanta et al. \cite{achanta2009frequency-tuned} to investigate the distribution of salient objects in photographs and then evaluate the influence on two state-of-the-art salient object detection models.

\section{Center Bias Model}\label{sec:hypotheses}

\begin{figure*}[tb]
  \centering
  \includegraphics[width=0.32\linewidth]{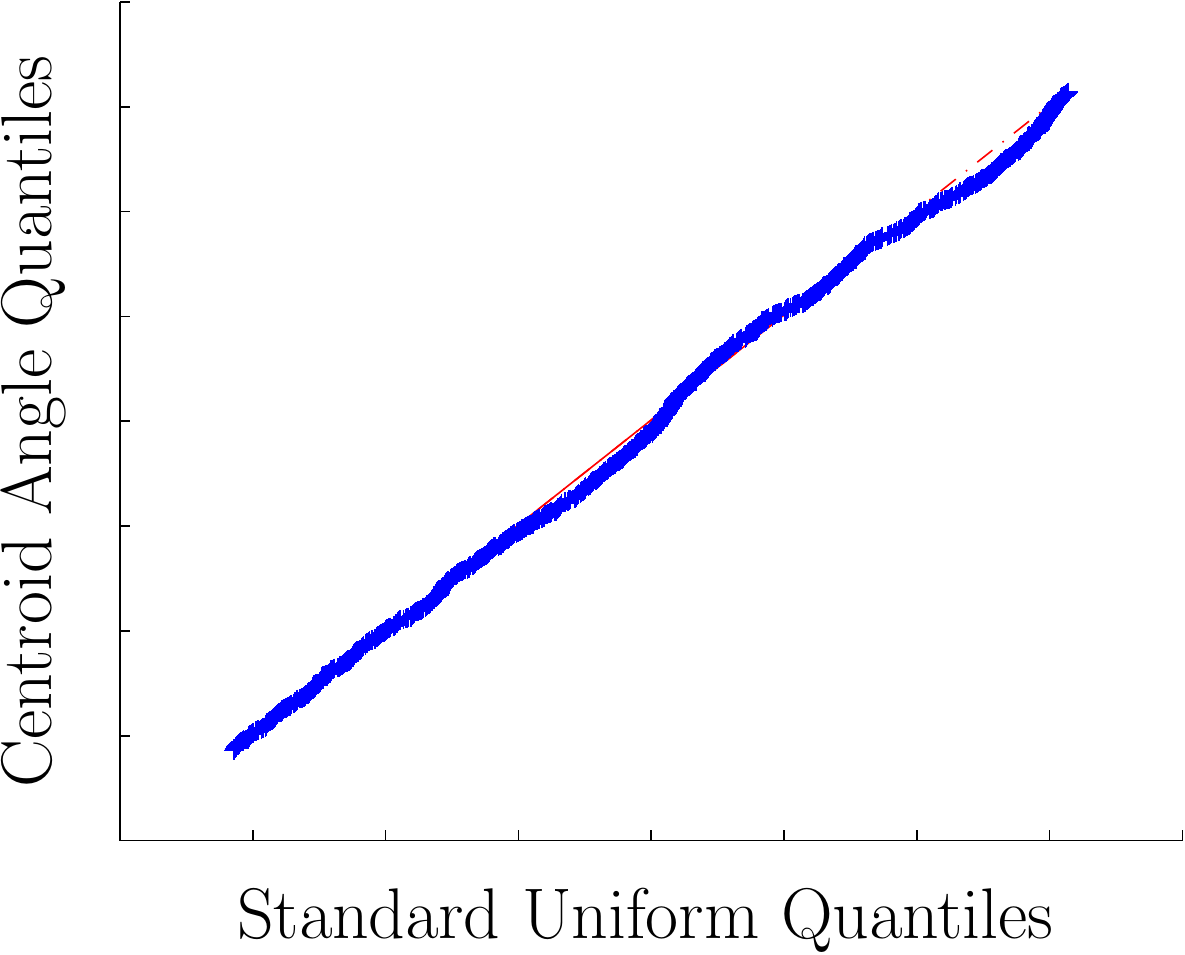}
  \hfill
  \includegraphics[width=0.32\linewidth]{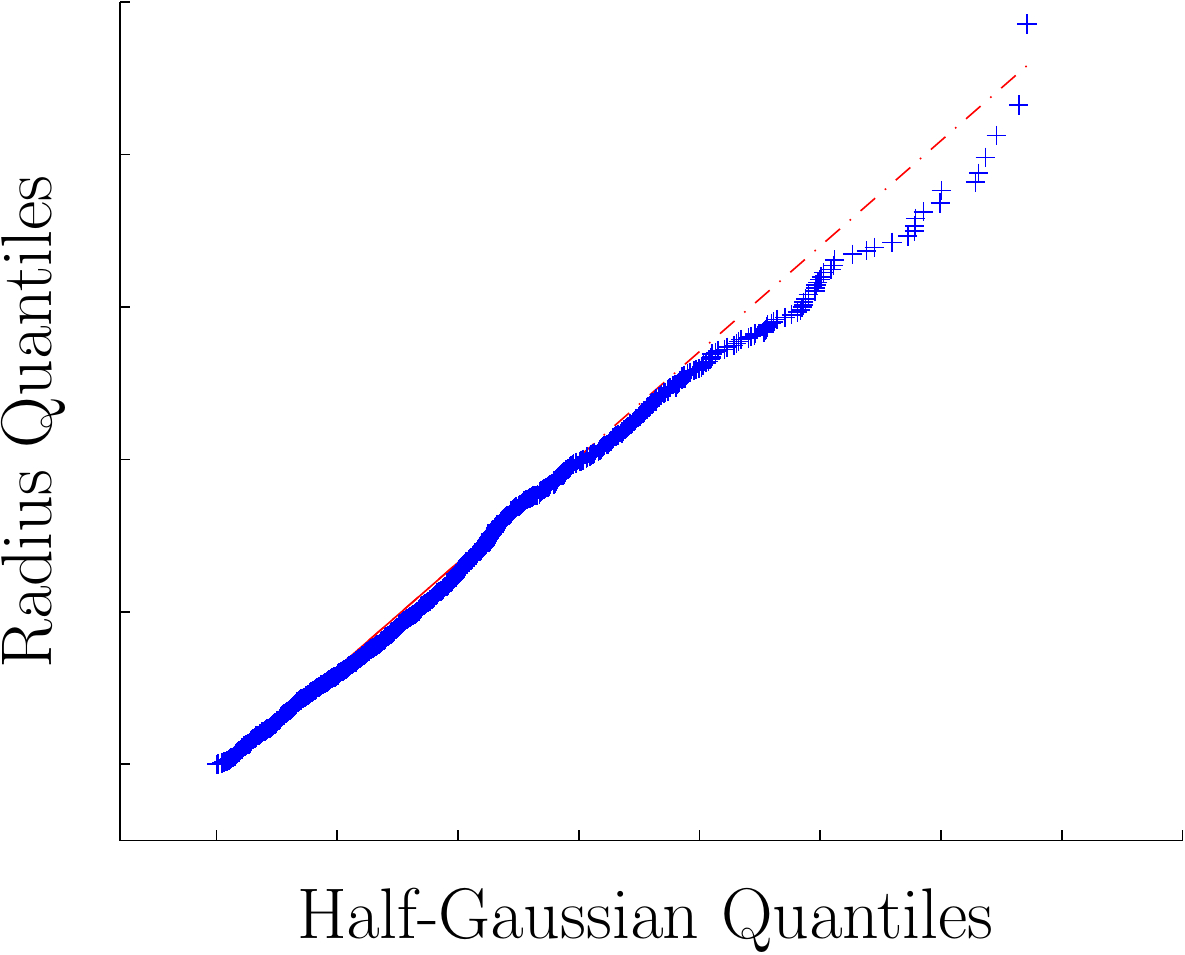}
  \hfill
  \includegraphics[width=0.32\linewidth]{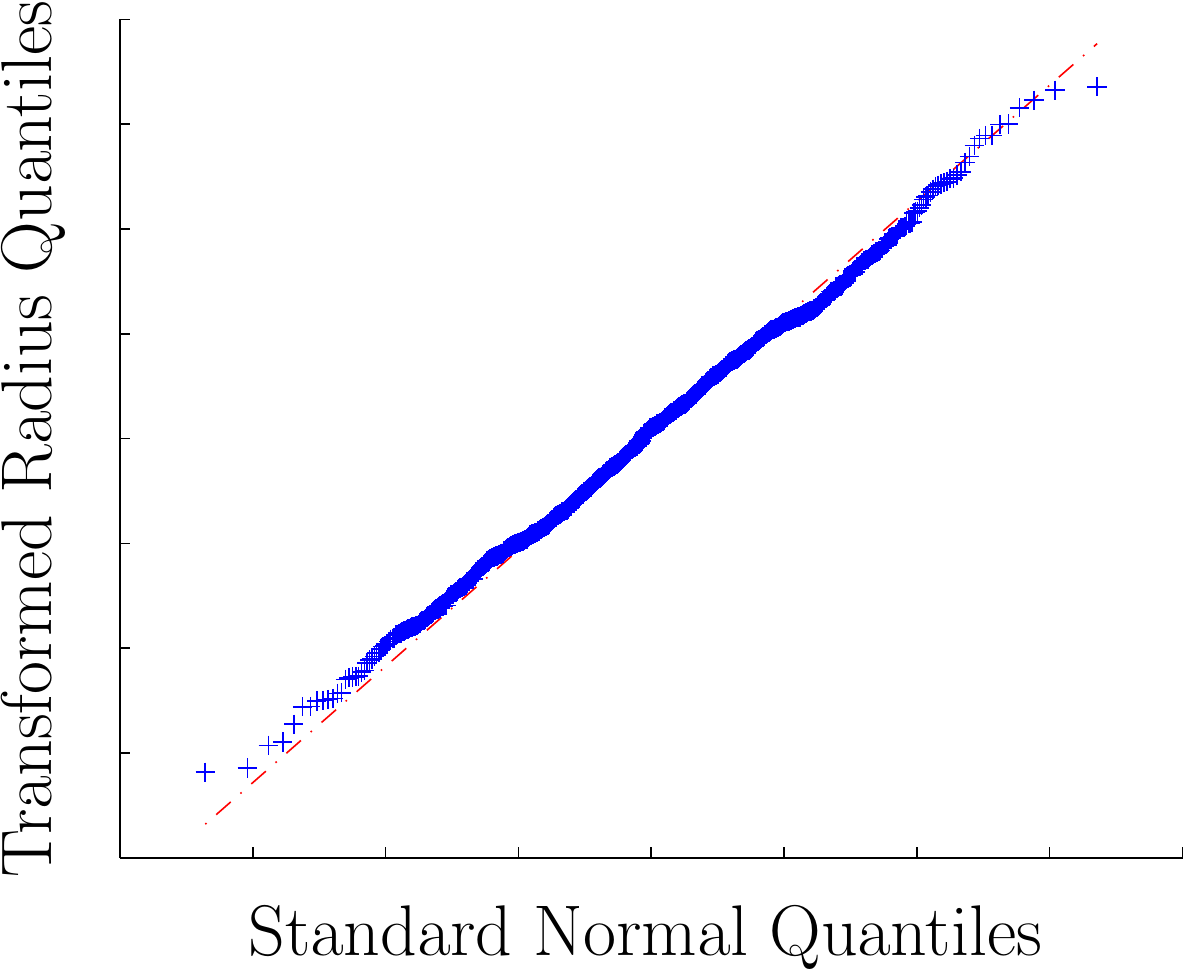}
  \caption{Quantile-Quantile (Q-Q) plots of the angles versus a uniform distribution (left), radii versus a half-Gaussian distribution (middle), transformed radii (see Sec.~\ref{sec:hypotheses:radius}) versus a normal distribution (right).}
  \label{fig:hypotheses:qqplots}
\end{figure*} 

To investigate the spatial distribution of salient objects in photographs collected from the web, we use the manually annotated segmentation masks by Achanta et al. \cite{achanta2009frequency-tuned,achanta2010saliency} that mark the salient objects in 1000 images of the salient object data set by Liu et al. \cite{liu2007learning}.
More specifically, we use the segmentation masks to determine the centroids of all salient objects in data set and analyze the centroids' spatial distribution.
The images in the data set by Liu et al. \cite{liu2007learning} have been collected from a variety of sources, mostly from image forums and image search engines.
Liu et al. collected more than 60,000 images and subsequently selected an image subset in which all images contain a salient object or a distinctive foreground object \cite{liu2007learning}.
9 users marked the salient objects using (rough) bounding boxes and the salient objects in the image database have been defined based on the \enquote{majority agreement}.
However, as a consequence of the selection process, the data set does not include images without distinct salient objects.
This is an important aspect to consider when trying to generalize the results reported on Achanta et al.'s and Liu et al.'s data set to other data sets or application areas.

In order to statistically analyze the 2-dimensional spatial distribution of the salient objects' centroids, we first identify the center of the spatial distribution.
Then, given the distribution's center, we can use a polar coordinate system to independently analyze the distribution of the angles and distances between the center and the salient objects.

\subsection{The Center}\label{sec:hypotheses:mode}

Our model is based on a polar coordinate system that has its pole at the image center.
Since the images in Achanta's data set have varying widths and heights, we use in the following normalized Cartesian 
image 
coordinates in the range $[0,\,1]\times[0,\,1]$.
The mean salient object centroid location is $[0.5021,\,0.5024]^{\mathrm{T}}$ and the corresponding covariance matrix is $\left[0.0223,\,-0.0008; -0.0008,\,0.0214\right]$. 
Thus, we can motivate the use of a polar coordinate system that has its pole at $[0.5,\,0.5]^{\mathrm{T}}$ to represent all locations relative to the expected distribution's mode.

\subsection{The Angles are Distributed Uniformly}\label{sec:hypotheses:angle}

Our first model hypothesis is that the centroids' angles in the specified polar coordinate system are uniformly distributed in $\left[-\pi,\pi\right]$.

In order to investigate the hypothesis, we use a Quantile-Quantile (Q-Q) plot as a graphical method to compare probability distributions (see \cite{nist2012handbook}).
In Q-Q plots the quantiles of the samples of two distributions are plotted against each other.
Thus, the more similar the two distributions are, the better the points in the Q-Q plot will approximate the line $f(x) = x$.
We calculate the Q-Q plot of the salient object location angles in our polar coordinate system versus uniformly drawn samples in $\left[-\pi,\pi\right]$, see Fig.~\ref{fig:hypotheses:qqplots} (left).
The apparent linearity of the plotted Q-Q points supports the hypothesis that the angles are distributed uniformly.

We can quantify the observed linearity, see Fig.~\ref{fig:hypotheses:qqplots} (left), to analyze the correlation between the model distribution and the data samples using probability plot correlation coefficients (PPCC) \cite{nist2012handbook}.
The PPCC is the correlation coefficient between the paired quantiles and measures the agreement of the fitted distribution with the observed data (i.e., goodness-of-fit).
The closer the correlation coefficient is to one, the higher the positive correlation and the more likely the distributions are shifted and/or scaled versions of each other.
Furthermore, by comparing against critical values of the PPCC (see \cite{vogel1989low-flow} and \cite{nist2012handbook}), we can use the PPCC as a statistical test, which is closely related to the Shapiro-Wilk test \cite{shapiro1965analysis} and can reject the hypothesis that the data samples match the assumed model distribution.
Furthermore, we can use the correlation to test the hypothesis of no correlation by transforming the correlation to create a t-statistic.

The obvious linearity of the Q-Q plot, see Fig.~\ref{fig:hypotheses:qqplots} (left), is reflected by a PPCC of $0.9988$\footnote{Mean of several runs with $N=1000$ uniform randomly selected samples.}, which is substantially higher than the critical value of $0.8880$ (see \cite{vogel1989low-flow}) and thus the hypothesis of identical distributions can not be rejected. 
Furthermore, the hypothesis of no correlation is rejected at $\alpha=0.05$ ($p = 0$).

\subsection{The Radii follow a Half-Gaussian Distribution}\label{sec:hypotheses:radius}

Our second model hypothesis is that the radii of the salient object locations follow a half-Gaussian distribution. 
We have to consider a half-Gaussian distribution in the interval $\left[0,\infty\right]$, because the radius -- as a length -- is by definition positive.
If we consider the image borders, we could assume a two-sided truncated distribution, but we have three reasons to work with a one-sided model:
The variance of the radii seems sufficiently small, the \enquote{true} centroid of the salient object may be outside the image borders (i.e., parts of the salient object can be truncated by the image borders), and it facilitates the use of various, well-known statistical tests (see \cite{schauerte2013distribution}).

We can use a Q-Q plot against a half-Gaussian distribution to graphically assess the hypothesis, see Fig.~\ref{fig:hypotheses:qqplots} (middle).
The linearity of the points suggests that the radii are distributed according to a half-Gaussian distribution. 
The visible outliers in the upper-right are caused by less than 30 centroids that are highly likely to be disturbed by the image borders.
Please be aware of the fact that it is not necessary to know the exact distribution parameters when working with Q-Q plots as long as the distributions are linearly related (see \cite{nist2012handbook}).
Furthermore, we transform the polar coordinates in such a way that they represent the same point with a combination of positive angles in $\left[0,\pi\right]$ and radii in $\left[-\infty,\infty\right]$.
This way, we can compare the distribution of the transformed radii against a normal distribution with its mode and mean at $0$, see Fig.~\ref{fig:hypotheses:qqplots} (right).

The obvious correlation that is visible in the Q-Q plots, see Fig.~\ref{fig:hypotheses:qqplots} (middle and right), is reflected by a PPCC of $0.9987$, which is above the critical value of $ 0.9984$ (see \cite{nist2012handbook}).
The hypothesis of no correlation is rejected at $\alpha=0.05$ ($p = 0$).

\section{Quantifying the Influence on Salient Object Detection}\label{sec:evaluation}

\label{sec:evaluation:salobj}

To assess the influence of the center bias on pixel- and object-based salient object detection, we integrate a Gaussian center bias into the algorithms by Achanta et al. \cite{achanta2010saliency} and Cheng at al. \cite{cheng2011global}.

\subsection{Center Biased Saliency Models}
\label{sec:model}

\newcommand*{\footref}[1]{\textsuperscript{\ref{#1}}}

\subsubsection{Pixel-based}\label{sec:model:pixel}

As a pixel-based model, we use maximum symmetric surround saliency detection by Achanta et al. \cite{achanta2010saliency} in combination with a Gaussian center bias map (cf., e.g., \cite{judd2009learning,borji2012probabilistic}).
To this end, we define the center bias saliency map $S_{\mathrm{C}} \in \mathbf{R}^{M \times N}$
\begin{align}
	S_{\mathrm{C}}(x,y) \quad=\quad& g(\mu_x - x,\mu_y - y;\sigma_x,\sigma_y) \quad \mathrm{with} \\
	g(x,y;\sigma_x,\sigma_y) \quad=\quad&   \frac{1}{\sqrt{2\pi}\sigma_x} \exp\left\{-\frac{1}{2}\frac{x^2}{\sigma_x^2}\right\} \label{eq:gaussian}\\  
                                & * \frac{1}{\sqrt{2\pi}\sigma_y} \exp\left\{-\frac{1}{2}\frac{y^2}{\sigma_y^2}\right\} ,\notag 
\end{align}
where $(x,y)$ is the pixel coordinate, $\mu = (\mu_x, \mu_y)$ is the image center's coordinate, and $\sigma_x$ and $\sigma_y$ are the standard deviation in x- and y-direction depending on the image width and height, respectively.

In order to investigate the influence of the center bias, we investigate different, plausible strategies to investigate the combination of the bottom-up and center bias saliency maps $S_{\mathrm{B}}$ and $S_{\mathrm{C}}$, respectively:
\begin{equation}
	S_{\mathrm{P}} = f(S_{\mathrm{C}}, S_{\mathrm{B}}) , \label{eq:combination:pixel:convex}
\end{equation}
where $f$ is the chosen center bias integration scheme.

We consider the following schemes, cf. \cite{schauerte2012predicting}:
First, a convex, linear integration, i.e. 
$f_{+}(w_{\mathrm{C}} S_{\mathrm{C}}, S_{\mathrm{B}}) = w_{\mathrm{C}} S_{\mathrm{C}} + w_{\mathrm{B}} S_{\mathrm{B}}$ with $w_{\mathrm{B}} + w_{\mathrm{C}} = 1$ ($w_{\mathrm{B}}, w_{\mathrm{C}} \in \mathbf{R}^+_0$).
Second, multiplicative integration as a supra-linear combination method, i.e. $f_{\circ}(w_{\mathrm{C}} S_{\mathrm{C}}, S_{\mathrm{B}}) = S_{\mathrm{C}} \circ S_{\mathrm{B}}$, where $\circ$ denotes the Hadamard product.
Third, the minimum as a further, alternative supra-linear combination, i.e. $f_{\downarrow}(w_{\mathrm{C}} S_{\mathrm{C}}, S_{\mathrm{B}}) = \min(S_{\mathrm{C}}, S_{\mathrm{B}})$.
Fourth, the maximum to realize a late, sub-linear combination scheme, i.e. $f_{\uparrow}(w_{\mathrm{C}} S_{\mathrm{C}}, S_{\mathrm{B}}) = \max(S_{\mathrm{C}}, S_{\mathrm{B}})$.
All these schemes are also related to different Fuzzy logic interpretations, 
which might provide a common theoretical framework and interpretation throughout later applications (e.g., \cite{schauerte2009multi-modal:icmi}).
To improve the readability, we refer to the linear combination for explicit center bias integration -- unless stated otherwise, of course -- in the following .

\subsubsection{Segmentation-based}\label{sec:model:segment}

\begin{figure}[tb]
  \centering
  \includegraphics[width=0.19\linewidth]{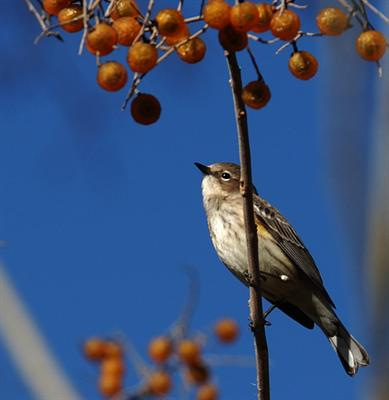}\hspace{1mm}
  \includegraphics[width=0.19\linewidth]{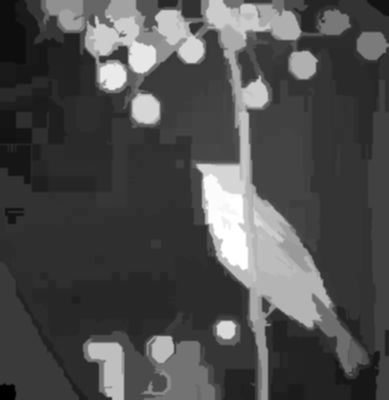}\hspace{1mm}
  \includegraphics[width=0.19\linewidth]{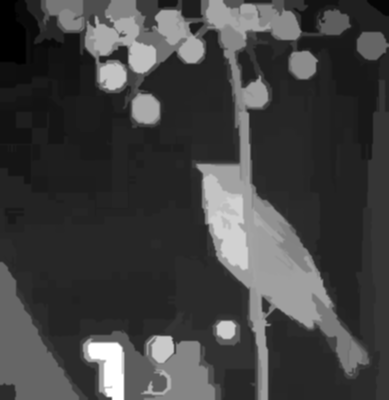}
  \caption{An example illustrating the influence of the implicit center bias in the region contrast method by Cheng et al. \cite{cheng2011global}. Left-to-right: Image, region contrast (w/o explicit center bias), and locally debiased region contrast (w/o explicit center bias).}
  \label{fig:model:example:implicit}
\end{figure}

\newcommand{\myparagraph}[1]{}

\begin{figure}[tb]
  \centering
  \includegraphics[width=0.19\linewidth]{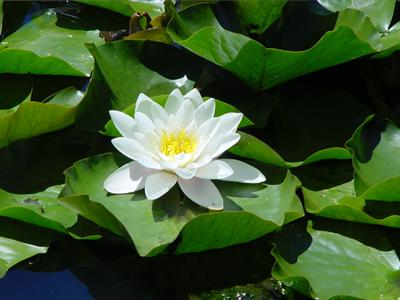}
  \includegraphics[width=0.19\linewidth]{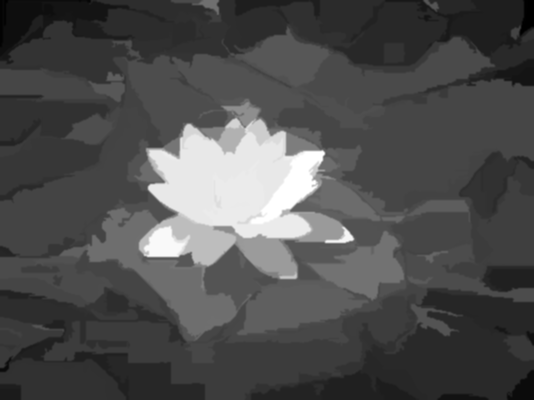}
  \includegraphics[width=0.19\linewidth]{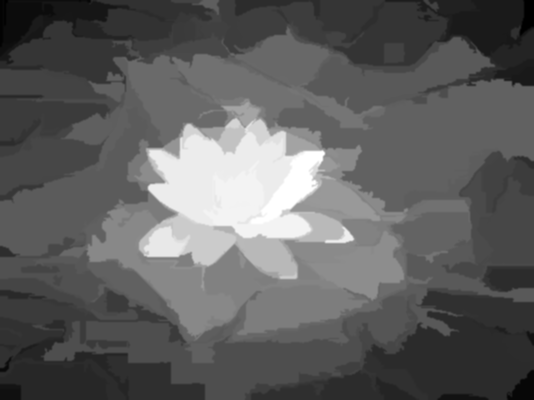}
  \includegraphics[width=0.19\linewidth]{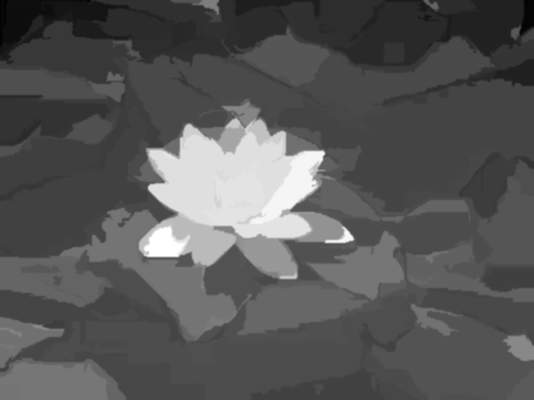}
  \includegraphics[width=0.19\linewidth]{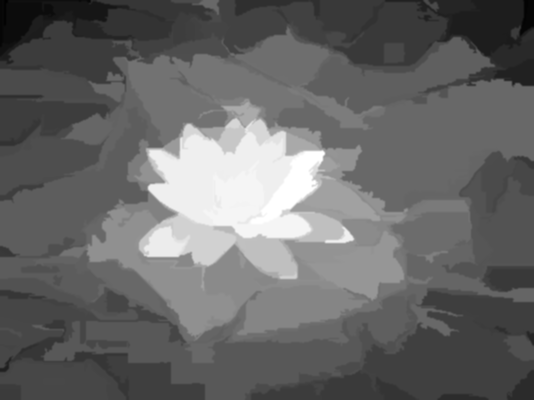}\vspace*{1mm}\\%
  \includegraphics[width=0.19\linewidth]{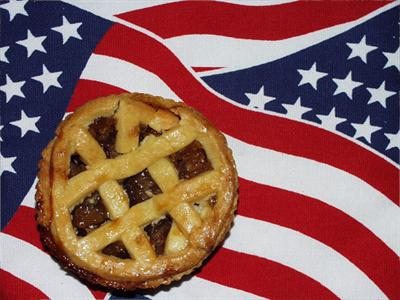}
  \includegraphics[width=0.19\linewidth]{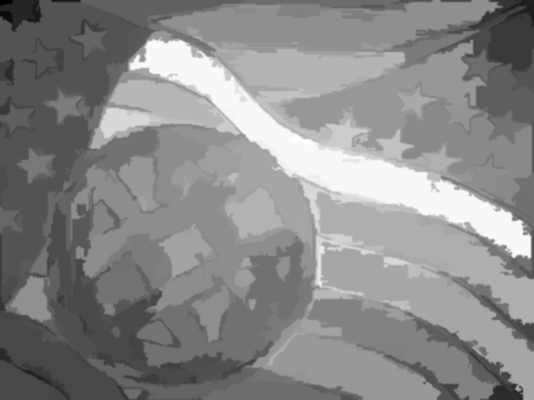}
  \includegraphics[width=0.19\linewidth]{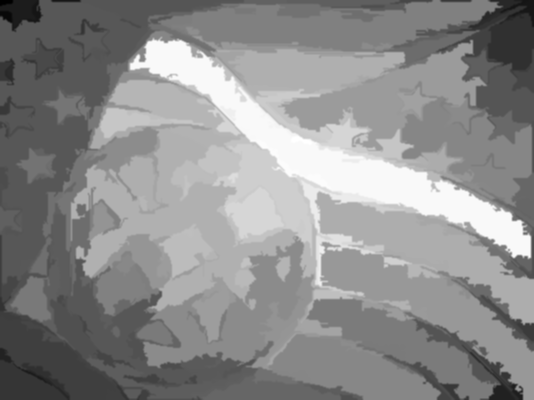}
  \includegraphics[width=0.19\linewidth]{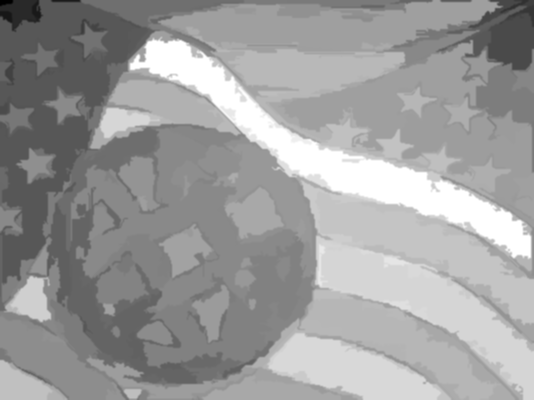}
  \includegraphics[width=0.19\linewidth]{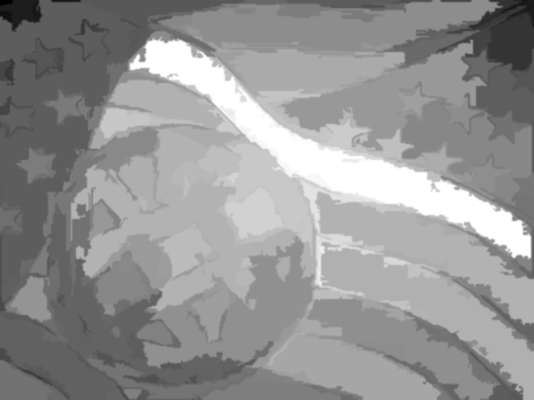}\vspace*{1mm}\\%
  \includegraphics[width=0.19\linewidth]{figs/examples/0_15_15264}
  \includegraphics[width=0.19\linewidth]{figs/examples/0_15_15264_smap_RC}
  \includegraphics[width=0.19\linewidth]{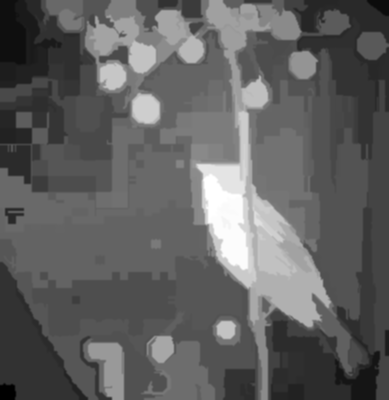}
  \includegraphics[width=0.19\linewidth]{figs/examples/0_15_15264_smap_LDRC}
  \includegraphics[width=0.19\linewidth]{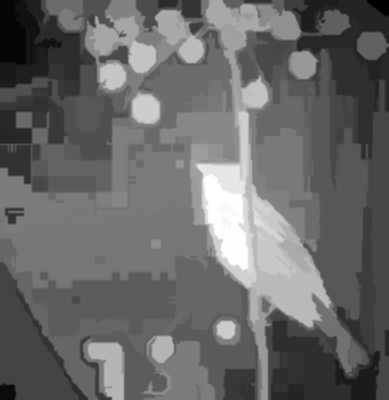}
  \caption{Examples of the influence of the implicit and explicit center bias on segmentation-based salient object detection. Left-to-right: Image, region contrast without and with center bias (RC and RC+CB, resp.), and locally debiased region contrast without and with center bias (LDRC and LDRC+CB, resp.).}
  \label{fig:model:example}

\end{figure}

\myparagraph{Explicit center bias term:} As a segmentation-based model, we adapt Cheng et al.'s region contrast model \cite{cheng2011global}. 
This model is particularly interesting, because it already provides state-of-the-art performance, which is partially caused by an implicit center bias as we will show in the following.
This way, we can observe how the model behaves if we remove the implicit center bias -- which was neither motivated nor explained by the authors -- and add an explicit Gaussian center bias.
The spatially weighted region contrast saliency equation is defined as follows
\begin{align}
    S_{\mathrm{S}}(r_k) =\quad &\sum_{r_k \neq r_i} \hat{D}_s(r_k;r_i)w(r_i)D_r(r_k; r_i) \qquad\text{with}\label{eq:model:segment}\\%\makebox[1cm][r]{with}\label{eq:model:segment}\\
    \hat{D}_s(r_k;r_i) =\quad &\exp(- D_s(r_k; r_i)/\sigma^2_s) \text{.}
\end{align}
$w(r_i)$ is the weight of region $r_i$, which equals the number of pixels in $r_i$ -- i.e., $w(r_i) = \card{r_i}$ -- to emphasize color contrast to bigger regions. 
$D_r(\cdot; \cdot)$ is the color distance metric between the two regions
\begin{align}
		D_r(r_1; r_2) = \sum_{c_1} \sum_{c_2} f(c_{1;i}) f(c_{2;j}) D(c_i; c_j) ,
\end{align}
where $f(c_{k;i})$ is the (frequentist) probability of the i-th color $c_{k;i}$ among all $n_k$ colors in the k-th region $r_k$, which is determined using a color histogram. 
The probability of the color inside the regions $f(c_{k;i})$ is used as weight to emphasize color differences between dominant colors.
$D(c_i; c_j)$ measures the distance between the colors and in the following it is defined as being the Euclidean distance in the CIE Lab color space.
Finally, 
$\hat{D}_s(r_k; r_i)$ is the spatial distance between regions $r_k$ and $r_i$, where $\sigma_s$ controls the spatial weighting.
The spatial distance between two regions is defined as the Euclidean distance between the centroids of the respective regions using pixel coordinates that are normalized to the range $[0,\,1] \times [0,\,1]$.
Smaller values of $\sigma_s$ influence the spatial weighting in such a way that the contrast to regions that are farther away contributes less to the saliency of the current region.

\begin{figure}[tb]
  \centering
  \includegraphics[width=0.20\linewidth,height=0.20\linewidth]{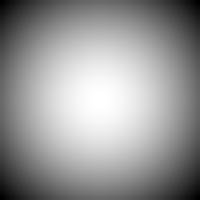}\label{fig:model:gaussianmetricbias:biased}\hspace{1mm}
  \includegraphics[width=0.20\linewidth,height=0.20\linewidth]{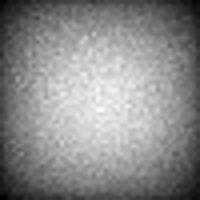}\label{fig:model:gaussianmetricbias:debiased}
  \caption{Illustration of the implicit center bias in the method by Cheng et al. \cite{cheng2011global}. Left: Each pixel shows the distance weight sum, i.e. $\sum_{r_i}\hat{D}_s(r_k;r_i)$, to all other pixels in a regular grid. Right: The average weight sum depending on the centroid location calculated on the Achanta/Liu data set using Felzenszwalb's segmentation method \cite{felzenszwalb2004efficient}.}
  \label{fig:model:gaussianmetricbias}

\end{figure}

It is this unnormalized Gaussian weighted Euclidean distance $\hat{D}_s(r_k; r_i)$ that causes an implicit Gaussian-like center bias (see Fig.~\ref{fig:model:example:implicit} and \ref{fig:model:gaussianmetricbias}), because it favors regions whose distances to the other neighbors are smaller, which is -- in general -- the case for segments at the center of the image.
Although this biased distance function has a significant impact on the performance, its choice has not been clearly motivated, discussed, or evaluated by Cheng et al.
To remove this implicit bias, we introduce a normalized, i.e. locally debiased, distance function $\check{D}_s(r_k;r_i)$ that still weights close-by regions higher than further away regions, but does not lead to an implicit center bias
\begin{align}
    \check{D}_s(r_k;r_i) &=& \frac{\hat{D}_s(r_k;r_i)}{\sum_{r_i}\hat{D}_s(r_k;r_i)} \mathrm{\ ,} \\ &\mathrm{i.e.\ } & \forall r_k : \sum_{r_i}\check{D}_s(r_k;r_i) = 1 .
\end{align}

Similar to the pixel-based model (see Sec.~\ref{sec:model:pixel}), we can now integrate an explicit center bias into the segmentation-based model
\begin{align}
    S_{\mathrm{S}}(r_k) =\quad &f\left(\sum_{r_k \neq r_i} \check{D}_s(r_k;r_i)w(r_i)D_r(r_k; r_i), \, g(C(r_k); \sigma_x,\sigma_y)\right) .
\end{align}
Here, $f$ is the chosen center bias integration function as in Eq.~\ref{eq:combination:pixel:convex}.
Furthermore, $C(r_k)$ denotes the centroid of region $r_k$ and $g$ is defined as in Eq.~\ref{eq:gaussian}.

\subsection{Quantitative Evaluation}

\subsubsection{Dataset}\label{sec:evaluation:dataset}

As for the graphical investigation of our hypotheses using Q-Q plots (see Fig.~\ref{fig:hypotheses:qqplots}), we use the manually annotated segmentation masks by Achanta et al. \cite{achanta2009frequency-tuned,achanta2010saliency}, see Sec.~\ref{sec:hypotheses}, to quantify the influence of the Gaussian center bias on salient object detection.

\subsubsection{Baseline Algorithms}\label{sec:evaluation:algorithms}

In order to compare our results, we use a set of 
saliency detection algorithms that we group into two coarse categories:
First, algorithms that were specifically proposed for salient object detection and, second, algorithms that have been proposed and evaluated in other contexts.
From the second category, we use:
The well-known saliency model by Itti and Koch \cite{itti1998model}, Graph-Based Visual Saliency (GBVS) by Harel at al. \cite{harel2007graph-based}, Context-Aware Saliency (CAS) by Goferman et al. \cite{goferman2010context-aware,goferman2012context-aware}, and the FFT's spectral residuals (FFT) and DCT image signatures (DCT) by Hou et al. \cite{hou2007saliency,hou2011image}.
For FFT and DCT, we optimized the resolution at which the saliency maps are calculated, which is the most important algorithm parameter and has a significant influence on the performance\footnote{We were surprised by the fact that the spectral approaches (i.e., FFT and DCT) performed so well, because the previously reported results for FFT stated otherwise (see, e.g., \cite{achanta2009frequency-tuned,klein2011center-surround,cheng2011global}). However, this can probably be explained by the fact that we analyzed the influence of the saliency map resolution on these approaches, which is their most important parameter and has a considerable influence on the results.}.
As baseline for salient object detection algorithms (first category), we use:
The Frequency-Tuned model (FT) by Achanta et al. \cite{achanta2009frequency-tuned}\footnote{When comparing with the results in \cite{achanta2009frequency-tuned}, please read the erratum that has been published at \url{http://ivrg.epfl.ch/supplementary_material/RK_CVPR09}}, the Bonn Information-Theoretic Saliency model (BITS) by Klein et al. \cite{klein2011center-surround}, the Maximum Symmetric Surround Saliency (MSSS) model by Achanta et al. \cite{achanta2010saliency}, and the Region Contrast (RC) model by Cheng et al. \cite{cheng2011global} that uses Felzenszwalb's image segmentation method \cite{felzenszwalb2004efficient}.
The latter two are the original algorithms we adapted.

Of course, we evaluate our adapted, center biased models: 
The maximum symmetric surround saliency with center bias (MSSS+CB; see Sec.~\ref{sec:model:pixel}) and the region contrast model with explicit center bias (RC+CB; see Sec.~\ref{sec:model:segment}).
In order to investigate the influence of the implicit center bias in the region contrast model (see Sec.~\ref{sec:model:segment}), we calculate the performance of the locally debiased region contrast model without and with explicit center bias (LDRC and LDRC+CB, respectively; see Sec.~\ref{sec:model:segment}).
Additionally, as a reference we provide the results for the standalone segment-based and pixel-based center bias models, i.e. $w_\mathrm{C} = 1$ (CB$_\mathrm{S}$ and CB$_\mathrm{P}$, respectively).

\begin{figure}[tb]
  \centering
  \includegraphics[width=0.475\linewidth]{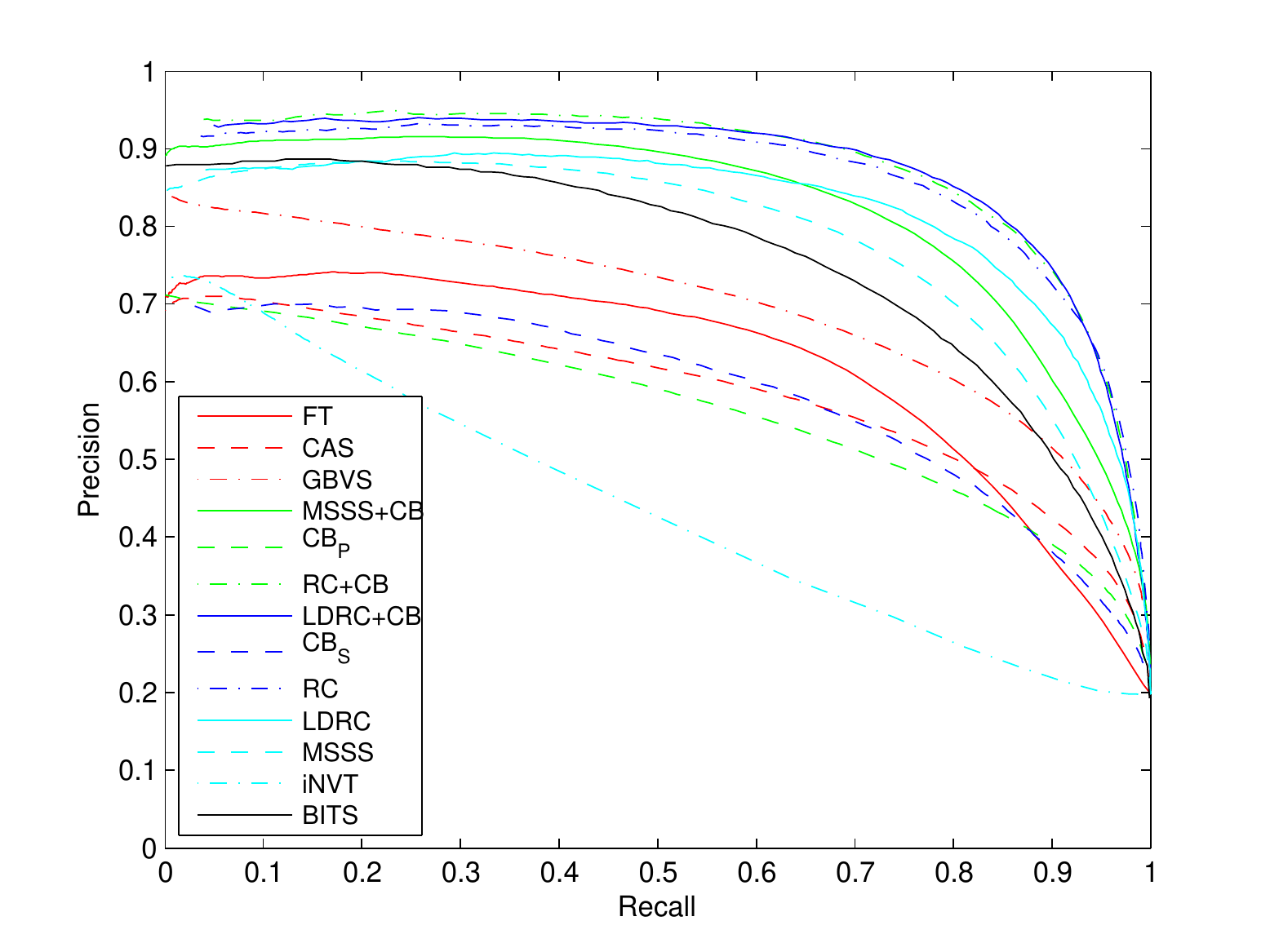}
  \includegraphics[width=0.475\linewidth]{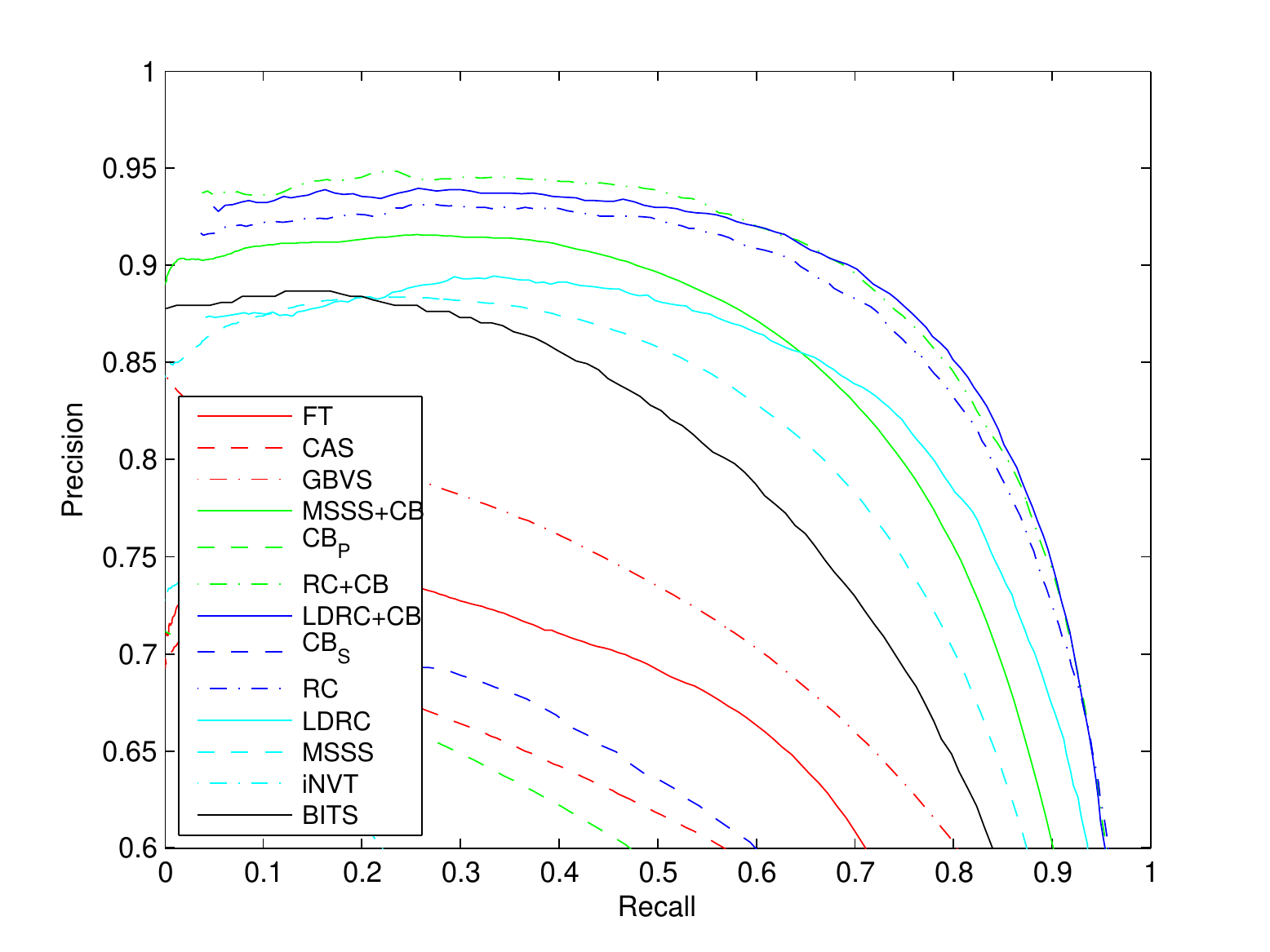}
  \caption{Precision-recall curves for all evaluated models with full (top) and limited range of the precision (bottom). This graphic is best viewed in color.}
  \label{fig:evaluation:pr}

\end{figure}

\paragraph{Implementation notes}
If available, we used the reference implementations that have been provided by the authors. 
For MSSS we use the C++ implementation by Achanta, because it provides a better performance than the basic Matlab implementation. 
For Itti we use the iLab Neuromorphic Vision Toolkit (iNVT). 
We integrated the methods directly into Matlab (mex) in order to avoid quantization and/or compression artifacts that may occur due to saving and loading them as images. 
For DCT and FFT, we used the implementations in our publicly available Matlab toolbox \cite{svist}. 
All calculations have been made using double precision arithmetic. 
To make our results as reproducible as possible (we have observed that the precision-recall curves of different authors vary), we will make our implementations and evaluation scripts open source.
We would like to note that our evaluation measure implementations follow the implementations of Weka and LingPipe. The corresponding precision-recall curves and results of further baseline algorithms can be seen in Fig.~\ref{fig:evaluation:pr}.

\subsubsection{Measures}\label{sec:evaluation:measures}

We can use the binary segmentation masks for saliency evaluation by treating the saliency maps as binary classifiers.
At a specific threshold $t$ we regard all pixels that have a saliency value above the thresholds as positives and all pixels with values below the thresholds as negatives.
By sweeping over all thresholds $\min(S) \leq t \leq \max(S)$, we can evaluate the performance using common binary classifier evaluation measures.

Most commonly, precision-recall curves are used -- e.g., by Achanta et al. \cite{achanta2010saliency,achanta2009frequency-tuned}, Cheng et al. \cite{cheng2011global}, and Klein et al. \cite{klein2011center-surround} -- to evaluate the salient object detection performance. 
We use five evaluation measures to quantify the performance of the algorithms.
We calculate the area under curve (AUC) of the (interpolated) precision-recall
curve (PR) and the receiver operating characteristic (ROC) curve \cite{davis2006relationship}.
Complementary to the PR AUC, we calculate the maximum $F_1$ and $F_{\sqrt{0.3}}$ scores with 
\begin{equation}
    F_\beta = (1 + \beta^2)\frac{\mathrm{precision}\cdot\mathrm{recall}}{\beta^2\cdot\mathrm{precision}+\mathrm{recall}} . 
\end{equation}
$F_{\beta}$ with $\beta=\sqrt{0.3}$ has been proposed by Achanta et al. to weight precision more than recall for salient object detection \cite{achanta2009frequency-tuned}.
Additionally, we calculate the hit-rate (HR) that measures how often the pixel with the maximum saliency belongs to the salient object.

\subsubsection{Results}\label{sec:evaluation:results}

\paragraph{Explicit center bias integration type}

How does the performance depend on the chosen center bias integration? To investigate this question, we tested the minimum, maximum, and product as alternative combinations. 
To account for the influence of different value distributions within the normalized value range, we also weighted the input of the $\min$ and $\max$ operation (e.g., $S^{\min}_\mathrm{P} = \min(w_\mathrm{C}S_\mathrm{C},w_\mathrm{B}S_\mathrm{B})$).
The results of the algorithms using different combination types are shown in Tab.~\ref{tab:evaluation:combinationtype}. 
The presented results are the results that we achieve 
with
the center bias weight that results in the highest $F_1$ score.

In Tab.~\ref{tab:evaluation:combinationtype}, we can see that the linear combination is the best choice for LDRC+CB.
However, for MSSS+CB and RC+CB the product seems to be the combination that provides the best performance.
Apparently MSSS+CB benefits more from using the product as combination type than RC+CB.
Also interesting to note is that LDRC+CB with the product as combination achieves similar results to RC.
However, LDRC+CB remains the algorithm that provides the best performance in terms of $F_1$ score and $F_\beta$ score whereas RC+CB provides the best performance in terms of PR AUC and HR.
Interestingly, LDRC+CB and RC+CB achieve a nearly identical ROC AUC.

\setlength{\tabcolsep}{5pt}
\begin{table*}[tb]
    \centering
    \begin{tabular}{l|l|ccccc}
        Method & Combination & $F_1$ & $F_\beta$ & $\int$PR & $\int$ROC & HR \\
        \hline
        LDRC+CB	& Linear/Convex & \ul{0.8034} & \ul{0.8183} &     0.8800  & \ul{0.9624} &     0.9240  \\
        LDRC+CB & Max           &     0.7504  &     0.7561  &     0.8108  &     0.9422  &     0.8630  \\
        LDRC+CB & Min           &     0.7897  &     0.8049  &     0.8584  &     0.9535  &     0.8880  \\
        LDRC+CB & Product       &     0.7883  &     0.8024  &     0.8704  &     0.9578  &     0.9130  \\
        RC+CB & Linear/Convex   &     0.7973  &     0.8120  &     0.8833  &     0.9620  &     0.9340  \\
        RC+CB & Max             &     0.7855  &     0.7993  &     0.8710  &     0.9568  &     0.9140  \\
        RC+CB & Min             &     0.7962  &     0.8150  &     0.8807  &     0.9603  &     0.9180  \\
        RC+CB & Product         &     0.7974  &     0.8136  & \ul{0.8878} & \ul{0.9623} & \ul{0.9460} \\
        MSSS+CB	& Linear/Convex &     0.7490  &     0.7678  &     0.8265  &     0.9495  &     0.8900  \\
        MSSS+CB & Max           &     0.7165  &     0.7337  &     0.7849  &     0.9270  &     0.8420  \\
        MSSS+CB & Min           &     0.7373  &     0.7606  &     0.8211  &     0.9339  &     0.9140  \\
        MSSS+CB & Product       &     0.7523  &     0.7748  &     0.8398  &     0.9445  &     0.9350  \\
        \hline
        LDRC	& --            &     0.7574  &     0.7675  &     0.8302  &     0.9430  &     0.8680  \\
        RC	    & --            &     0.7855  &     0.7993  &     0.8710  &     0.9568  &     0.9140  \\
        MSSS	& --            &     0.7165  &     0.7337  &     0.7849  &     0.9270  &     0.8420  \\
        \hline
        CB$_\mathrm{S}$	& --    &     0.5793  &     0.5764  &     0.5920  &     0.8623  &     0.6980  \\
        CB$_\mathrm{P}$	& --    &     0.5604  &     0.5452  &     0.5638  &     0.8673  &     0.7120  \\
    \end{tabular}
    \caption{The maximum $F_1$ score, maximum $F_\beta$ score, PR AUC ($\int$PR), ROC AUC ($\int$ROC), and Hit-Rate (HR) that we obtain using different combination types.\label{tab:evaluation:combinationtype}}
\end{table*}

\begin{figure*}[bt]
  \centering
  \subfigure[RC]{\includegraphics[trim=1.5cm 1cm 1.5cm 2.5cm, width=0.31\linewidth]{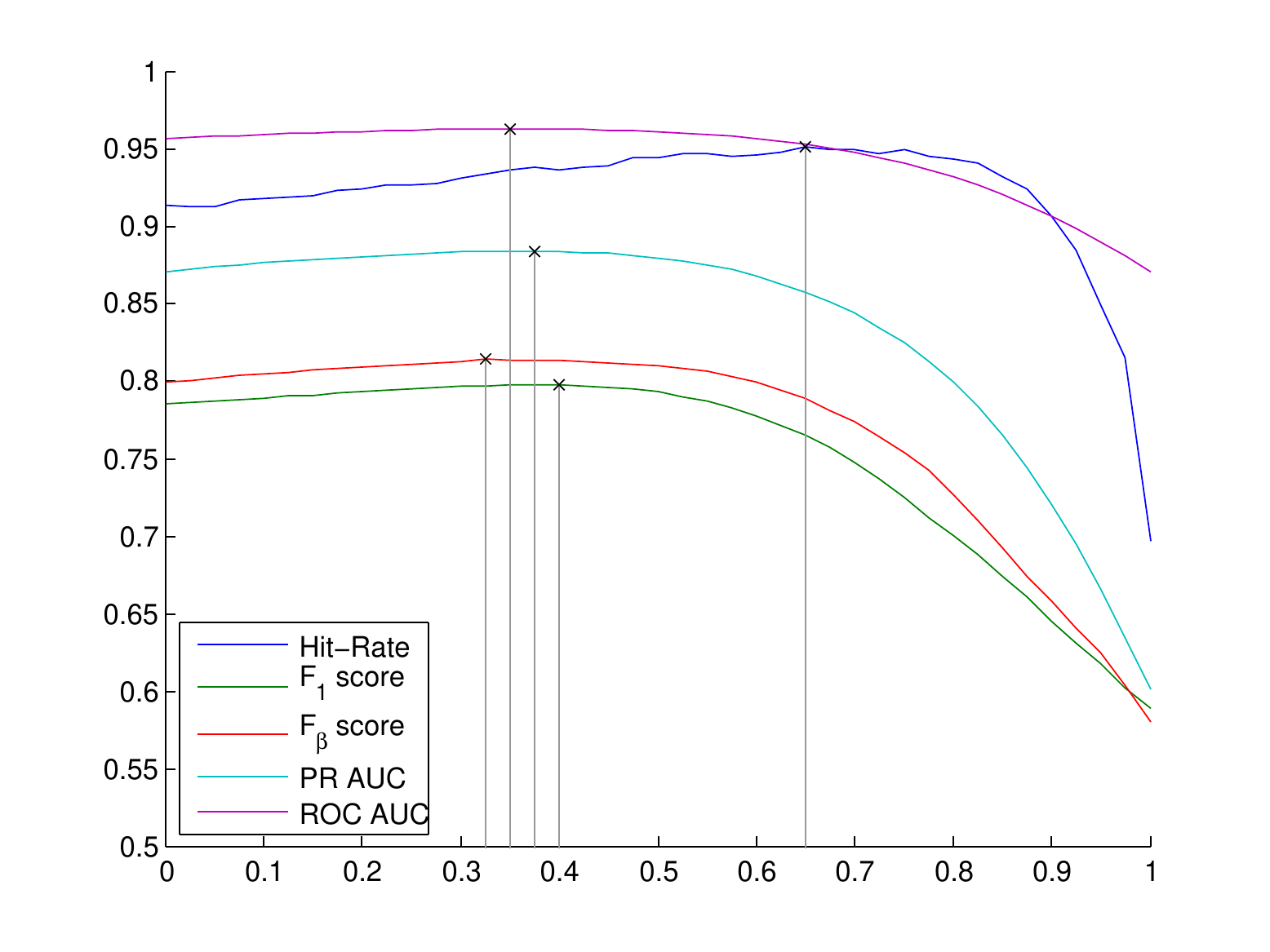}\label{fig:evaluation:combinationtype:ldrccb:linear}}\hspace{0.1cm}
  \subfigure[MSSS]{\includegraphics[trim=1.5cm 1cm 1.5cm 2.5cm, width=0.31\linewidth]{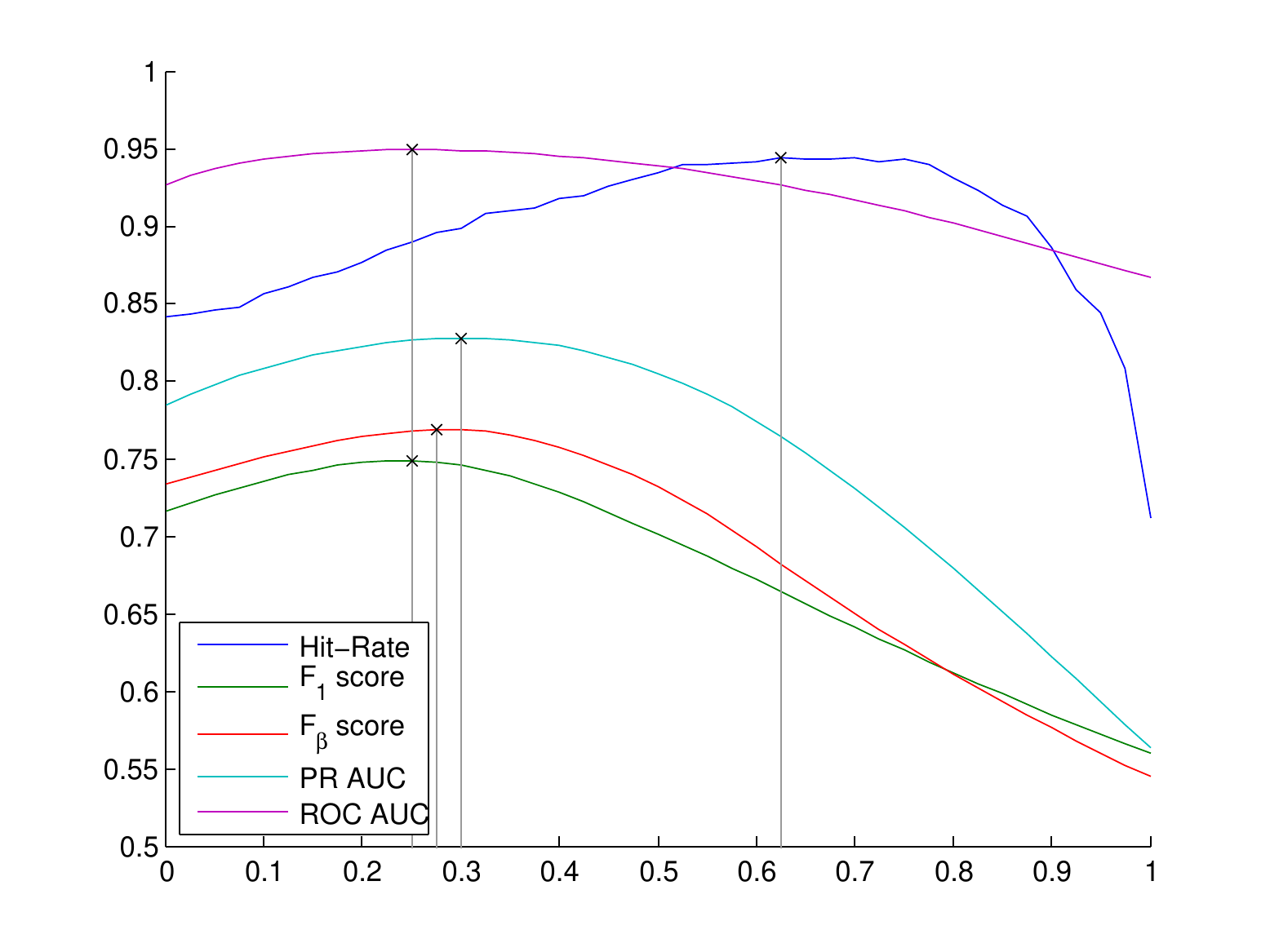}\label{fig:evaluation:combinationtype:rccb:linear}}\hspace{0.1cm}
  \subfigure[LDRC]{\includegraphics[trim=1.5cm 1cm 1.5cm 2.5cm, width=0.31\linewidth]{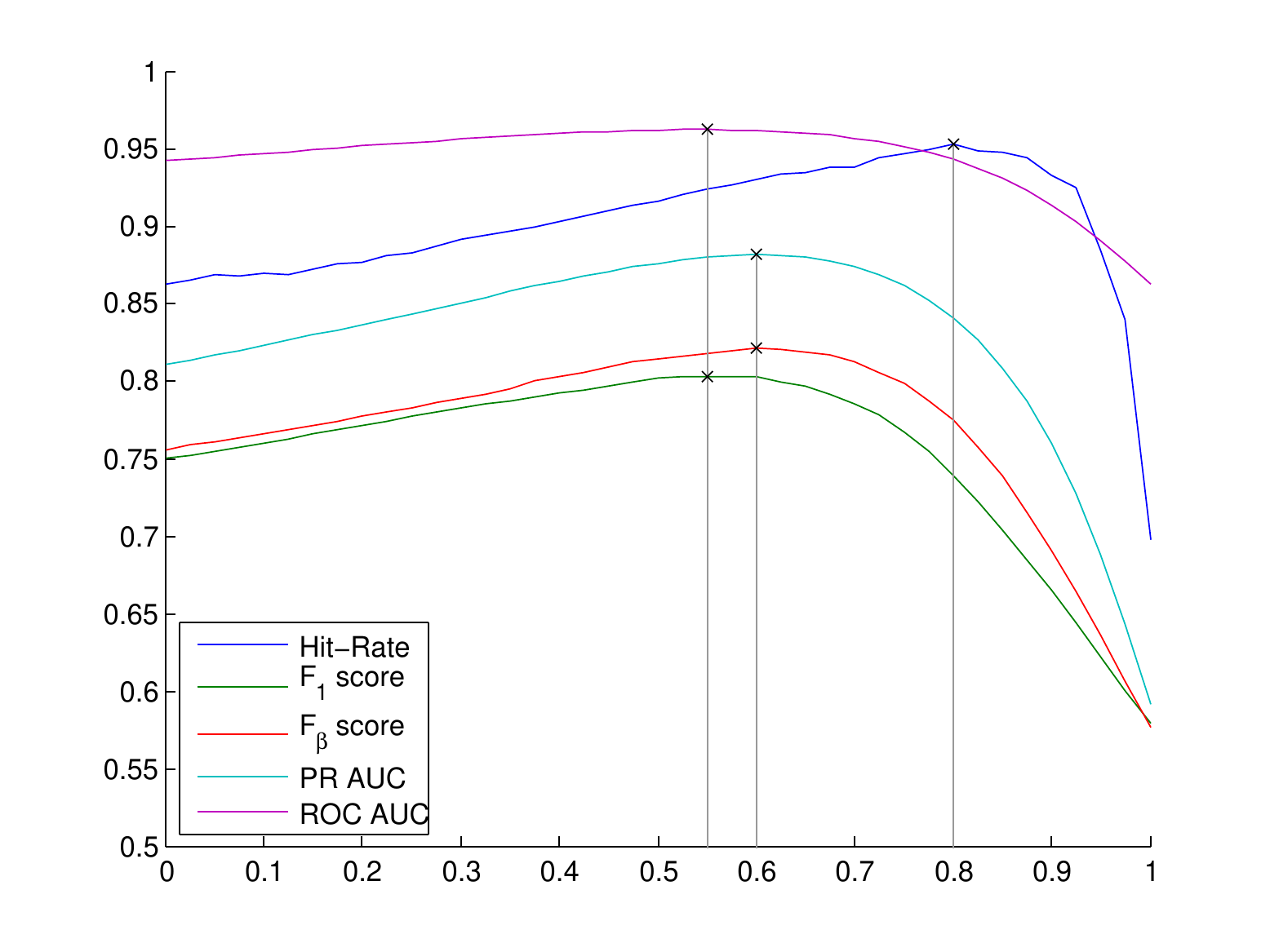}\label{fig:evaluation:combinationtype:mssscb:linear}}
  \caption{Illustration of the influence of the weight $w_\mathrm{C}$ on the performance of RC+CB, LDRC+CB, and MSSS+CB (convex combination).}  
\end{figure*}

\paragraph{Convex center bias weight}
How does the weight of the center bias influence the performance? To answer this question, we calculated the performance of LDRC+CB, RC+RB, and MSSS+CB with $w_\mathrm{C} \in \left[0,1\right]$ in $0.025$ steps. The resulting curves of the $F_1$ score, $F_\beta$ score, PR AUC, ROC AUC, and hit-rate are shown in Fig.~\ref{fig:evaluation:combinationtype:ldrccb:linear}, \ref{fig:evaluation:combinationtype:rccb:linear} and \ref{fig:evaluation:combinationtype:mssscb:linear}, respectively.

For each of the three algorithms the values of $w_\mathrm{C}$ that lead to the optimal $F_1$ score, $F_\beta$ score, PR AUC, and ROC AUC lie within a small interval.
In contrast, for all algorithms the value of $w_\mathrm{C}$ that achieves the highest hit-rate is outside these intervals and substantially higher.
Furthermore, the best weight for each measure depends on the algorithm and varies substantially.
It is interesting to see that small weights only have a minor (yet positive) influence on RC+CB until a point is reached (roughly at $w_\mathrm{C}=0.55$) where the performance begins to drop significantly.
This becomes especially apparent when comparing the curves of RC+CB, see Fig.~\ref{fig:evaluation:combinationtype:rccb:linear}, with the curves of LDRC+CB, see Fig.~\ref{fig:evaluation:combinationtype:ldrccb:linear}.

\paragraph{Quantitative comparison}

The center bias itself already has a considerable predictive power, see Tab.~\ref{tab:evaluation:results}, and is relatively close to the performance of FT. 
However, there is a substantial performance gap between the standalone center bias models (CB$_\mathrm{S}$ and CB$_\mathrm{P}$) and good non-biased methods such as, e.g., MSSS and LDRC.

\setlength{\tabcolsep}{5pt}
\begin{table}[tb]
    \centering
    \begin{tabular}{l|ccccc}
        Method & $F_1$ & $F_\beta$ & $\int$PR & $\int$ROC & HR \\
        \hline
        LDRC+CB	 & \ul{0.8034} & \ul{0.8183} &     0.8800  & \ul{0.9624} &     0.9240  \\
        RC+CB	 &     0.7973  &     0.8120  & \ul{0.8833} &     0.9620  & \ul{0.9340} \\
        RC	     &     0.7855  &     0.7993  &     0.8710  &     0.9568  &     0.9140  \\
        MSSS+CB	 &     0.7490  &     0.7678  &     0.8265  &     0.9495  &     0.8900  \\
        LDRC	 &     0.7574  &     0.7675  &     0.8302  &     0.9430  &     0.8680  \\
        BITS	 &     0.7342  &     0.7582  &     0.7589  &     0.9316  &     0.7540  \\
        MSSS	 &     0.7165  &     0.7337  &     0.7849  &     0.9270  &     0.8420  \\
        FFT	     &     0.6455  &     0.6375  &     0.6593  &     0.8926  &     0.8080  \\
        DCT	     &     0.6472  &     0.6368  &     0.6612  &     0.8962  &     0.8270  \\
        GBVS	 &     0.6403  &     0.6242  &     0.6970  &     0.9088  &     0.8480  \\
        FT	     &     0.5995  &     0.6009  &     0.6261  &     0.8392  &     0.7100  \\
        CB$_\mathrm{S}$	 & 0.5793 & 0.5764 & 0.5920 & 0.8623 & 0.6980 \\
        CAS	     & 0.5857 & 0.5615 & 0.5888 & 0.8741 & 0.6920 \\
        CB$_\mathrm{P}$	 & 0.5604 & 0.5452 & 0.5638 & 0.8673 & 0.7120 \\
        iNVT	 & 0.3383 & 0.4012 & 0.4396 & 0.5768 & 0.6870 \\

    \end{tabular}
    \caption{The maximum $F_1$ score, maximum $F_\beta$ score, PR AUC ($\int$PR), ROC AUC ($\int$ROC), and Hit-Rate (HR) of the evaluated algorithms (sorted ascending by $F_\beta$).\label{tab:evaluation:results}}
\end{table}

\setlength{\tabcolsep}{3pt}
\begin{table}[tb]
  \centering
  \begin{tabular}{l|l|ccccc}
    Method & Baseline & $F_1$ & $F_\beta$ & $\int$PR & $\int$ROC & HR \\
    \hline
    LDRC & RC & 96.4  & 96.0  & 95.3  & 98.6  & 95.0 \\
    \hline
    RC+CB & RC & 101.5  & 101.6  & 101.4  & 100.5  & 102.2 \\
    LDRC+CB & RC & 102.3  & 102.4  & 101.0  & 100.6  & 101.1 \\
    \hline
    LDRC+CB & LDRC & 106.1  & 106.6  & 106.0  & 102.1  & 106.5 \\
    MSSS+CB & MSSS & 104.5  & 104.7  & 105.3  & 102.4  & 105.7 \\
  \end{tabular}
  \caption{Relative performance (in \%) of our adapted algorithms with respect to their baseline.\label{tab:evaluation:relative}}
\end{table}

As could be expected, the performance of RC drops substantially if we remove the implicit center bias as is done by LDRC (see Sec.~\ref{sec:model:segment}), which can best be seen in Tab.~\ref{tab:evaluation:relative}.
What happens if we add our explicit center bias model to unbiased models?
As can be seen in the performance difference between MSSS and MSSS+CB as well as the performance difference between LDRC to LDRC+CB, the performance is substantially increased with respect to all evaluation measures, see Tab.~\ref{tab:evaluation:results} and \ref{tab:evaluation:relative}.
Interestingly, the relative performance improvement from pixel-based MSSS to MSSS+CB and segment-based LDRC to LDRC+CB is comparable, see Tab.~\ref{tab:evaluation:relative}.
Furthermore, with the exception of HR, the performance of LDRC+CB and RC+CB is nearly identical with a slight advantage for LDRC+CB (see Tab.~\ref{tab:evaluation:results} and Tab.~\ref{tab:evaluation:relative}).
This indicates that we did not lose important information by debiasing the distance metric (LDRC+CB vs RC+CB) and that the explicit Gaussian center bias model is advantageous compared to the implicit weight bias (LDRC+CB and RC+CB vs RC).

In summary, MSSS+CB provides a substantially higher performance than MSSS and outperforms, e.g., FT and BITS.
RC+CB and LDRC+CB provide a better performance than their unbiased counterparts RC and LDRC, respectively. 
Furthermore, their performance is very similar and both outperform all other models.
Interestingly, LDRC is the best model without center bias in our evaluation on Achanta's data set.
This makes LDRC an interesting candidate for applications in which the image data can not be expected to have a photographer's center bias (e.g., image data of surveillance cameras, autonomous robots, or human-robot interaction \cite{schauerte2014look}).

\paragraph{Statistical significance}
One question remains: 
Does the integration of an explicit center bias result in a statistically significant performance improvement?
To address this question, we test the performance (i.e., $F_1$, $F_\beta$, $\int$PR, and $\int$ROC) of LDRC and MSSS with and without an explicit center bias.
For this purpose, we rely on two pairwise, two-sample t-tests:
First, we perform a two-tailed test to check whether the compared performances with and without an integrated center bias come from distributions with equal means (i.e., $\mathcal{H}_{=}$: \enquote{means are equal}).
Second, we perform a one-tailed test to check whether the performance with an integrated center bias is worse that without an integrated center bias, i.e. the center biased performance distribution's mode is lower (i.e., $\mathcal{H}_{<}$: \enquote{mean is lower}).
If we can reject both hypotheses, then it is clear that the performance of the algorithm has significantly improved due to the integrated center bias.
All tests are performed at a confidence level of $95\%$, i.e., $\alpha = 5\%$.

For MSSS, we can reject the hypothesis of equal mean for $F_1$, $F_\beta$, $\int$PR, and $\int$ROC with $p_{F_1} = 0.0285$, $p_{F_\beta} = 0.0031$, $p_{\int\text{PR}} = 5.252\times10^{-7}$, and $p_{\int\text{ROC}} = 2.618\times10^{-16}$, respectively.
Additionally, we can reject the hypothesis that an integrated center bias has a negative influence on the performance with $p_{F_1} = 0.0142$, $p_{F_\beta} = 0.0015$, $p_{\int\text{PR}} = 2.626\times10^{-7}$, and $p_{\int\text{ROC}} = 1.309\times10^{-16}$.

Similarly, we can reject the hypothesis that the performance of LDRC with and without center bias has an equal mean for $F_1$, $F_\beta$, $\int$PR, and $\int$ROC with $p_{F_1} = 0.0018$, $p_{F_\beta} = 2.426\times10^{-5}$, $p_{\int\text{PR}} = 1.118\times10^{-7}$, and $p_{\int\text{ROC}} = 1.555\times10^{-5}$, respectively.
And, we can reject the hypothesis that an integrated center bias has a negative influence on the performance with $p_{F_1} = 9.071\times10^{-4}$, $p_{F_\beta} = 1.213\times10^{-5}$, $p_{\int\text{PR}} = 5.590\times10^{-8}$, and $p_{\int\text{ROC}} = 7.773\times10^{-6}$.

Consequently, it is apparent that the integration of a center bias can lead to statistically significant performance improvements for pixel-based as well as segmentation-based algorithms.
\section{Conclusion}\label{sec:conclusion}

We formulated and investigated two hypotheses about the location of salient objects in photographs: 
First, the radial centroid distribution around the image center is uniform.
Second, the distances between their centroids and the image center follow a normal distribution.
We investigated these hypotheses using graphical methods, which indicate that our hypotheses are true.
This is an important insight, because it provides a strong empirical motivation and justification for the widely applied Gaussian center bias models.
To investigate the influence of the center bias on salient object detection, we explicitly integrated the center bias model in two state-of-the-art salient object detection algorithms.
We have shown that the explicitly modeled center bias has a significant, positive influence on the performance (in terms of hit-rate, the area under the precision-recall curve, the area under the receiver operating characteristic curve, the $F_1$ score, and the $F_\beta$ score).
Last but not least, by debiasing Cheng et al.'s region contrast model, we have exemplarily shown that implicit center biases might at least partially be responsible for the performance of state-of-the-art salient object detection algorithms and as a consequence we introduced an adapted, non-biased salient object detection algorithm.

% \section*{Acknowledgments}
% The work presented in this paper was supported by the German Research Foundation (DFG) within the Collaborative Research Program SFB 588 \enquote{Humanoide Roboter} and the Quaero Programme, funded by OSEO, French State agency for innovation.

\bibliographystyle{model1-num-names}
\bibliography{IEEEabrv,bibs/abrv,bibs/main}

\end{document}